\PassOptionsToPackage{table}{xcolor}
\documentclass[letterpaper, preprint]{article}
\usepackage[]{aaai2027}
\usepackage[hyphens]{url}
\usepackage{graphicx}
\usepackage{amsmath}
\usepackage{natbib}
\usepackage{caption}
\usepackage{booktabs}
\usepackage{amssymb}
\usepackage{multirow}
\usepackage{algorithm}
\usepackage{algorithmic}
\usepackage{xcolor}
\usepackage{tabularx}
\definecolor{ReferenceRow}{HTML}{F4F5F5}
\definecolor{OursRow}{HTML}{EAF5F1}

\newtheorem{proposition}{Proposition}
\usepackage{booktabs}
\usepackage{multirow}
\usepackage{array}
\usepackage{siunitx}

\title{Approximate Speculative Decoding}
\author{
Yuannuo Feng\textsuperscript{\rm 1\equalcontrib},
Zegang Peng\textsuperscript{\rm 2\equalcontrib}
Yuxin Xie\textsuperscript{\rm 1},
Yubing Ye\textsuperscript{\rm 3},
Yizhe Chen\textsuperscript{\rm 1},\\
Wenshuai Yao\textsuperscript{\rm 5},
Wenyong Zhou\textsuperscript{\rm 4}\corresponding,
Wang Kang\textsuperscript{\rm 1}\corresponding
}
\affiliations{
\textsuperscript{\rm 1}School of Integrated Circuit Science and Engineering, Beihang University, Beijing, China\\
\textsuperscript{\rm 2}Department of Precision Instrument, Tsinghua University, Beijing, China\\
\textsuperscript{\rm 3}Faculty of Engineering, The University of Hong Kong, Hong Kong SAR, China\\
\textsuperscript{\rm 4}Department of Electrical and Computer Engineering, The University of Hong Kong, Hong Kong SAR, China\\

\textsuperscript{\rm 5}School of Integrated Circuits, Peking University, Beijing, China
}

\begin{document}

\maketitle

\begin{abstract}
Speculative decoding accelerates autoregressive generation by verifying a draft block with a target model in parallel. Under standard greedy verification, decoding stops at the first draft token that differs from the target argmax, discarding the remaining target-scored suffix. Although accepting such a mismatch changes the decoding trajectory, it can make a contiguous suffix reusable when its tokens remain target-greedy under the realized prefix.
In this paper, we introduce \textbf{Approximate Speculative Decoding (ASD)}, a training-free verifier that replaces binary first-mismatch truncation with budgeted longest-prefix selection. ASD accepts selected mismatches subject to a local target-logit regret gate, a per-block exception cap, and a persistent request-level regret budget, then reuses the contiguous target-greedy suffix without additional approximate decisions or target-model forward passes. ASD requires neither a new draft model nor fine-tuning, and exactly reduces to standard greedy verification when the budget is zero.
Experiments show that ASD improves fixed-workload throughput by $3.05\%$--$15.26\%$ over matched strict verification and averages a $7.78\%$ gain across seven Qwen3-14B + DSpark-14B tasks. On DeepSeek-V4-Flash (284B) with DSpark it also raises verifier-side acceptance by roughly $10\%$--$16\%$ on GSM8K and MATH-500 in an FP4-to-FP8 compatibility setting. The source code is publicly available at: https://github.com/Kissmetothemoon/ASD
\end{abstract}

\section{Introduction}

Autoregressive large language models (LLMs) generate tokens sequentially, requiring repeated execution of a large target model and making decoding a major inference bottleneck. Early blockwise decoding exposed parallelism by predicting several future tokens at once~\cite{stern2018blockwise}; speculative decoding subsequently paired a lightweight proposal model with parallel target verification while preserving the target behavior~\cite{leviathan2023fast,chen2023accelerating}. This framework has since expanded across learned drafters, self-speculation, and structured draft trees, making both proposal quality and verification policy central to practical acceleration. Under standard greedy verification, the verifier commits the longest prefix whose draft tokens exactly match the target argmax, stops at the first mismatch, and appends a target recovery token. This rule exactly preserves the target greedy trajectory, but it is inherently binary: a nearly tied draft token is treated identically to a strongly disfavored one.
\begin{figure}[!t]
\centering
\includegraphics[width=\linewidth]{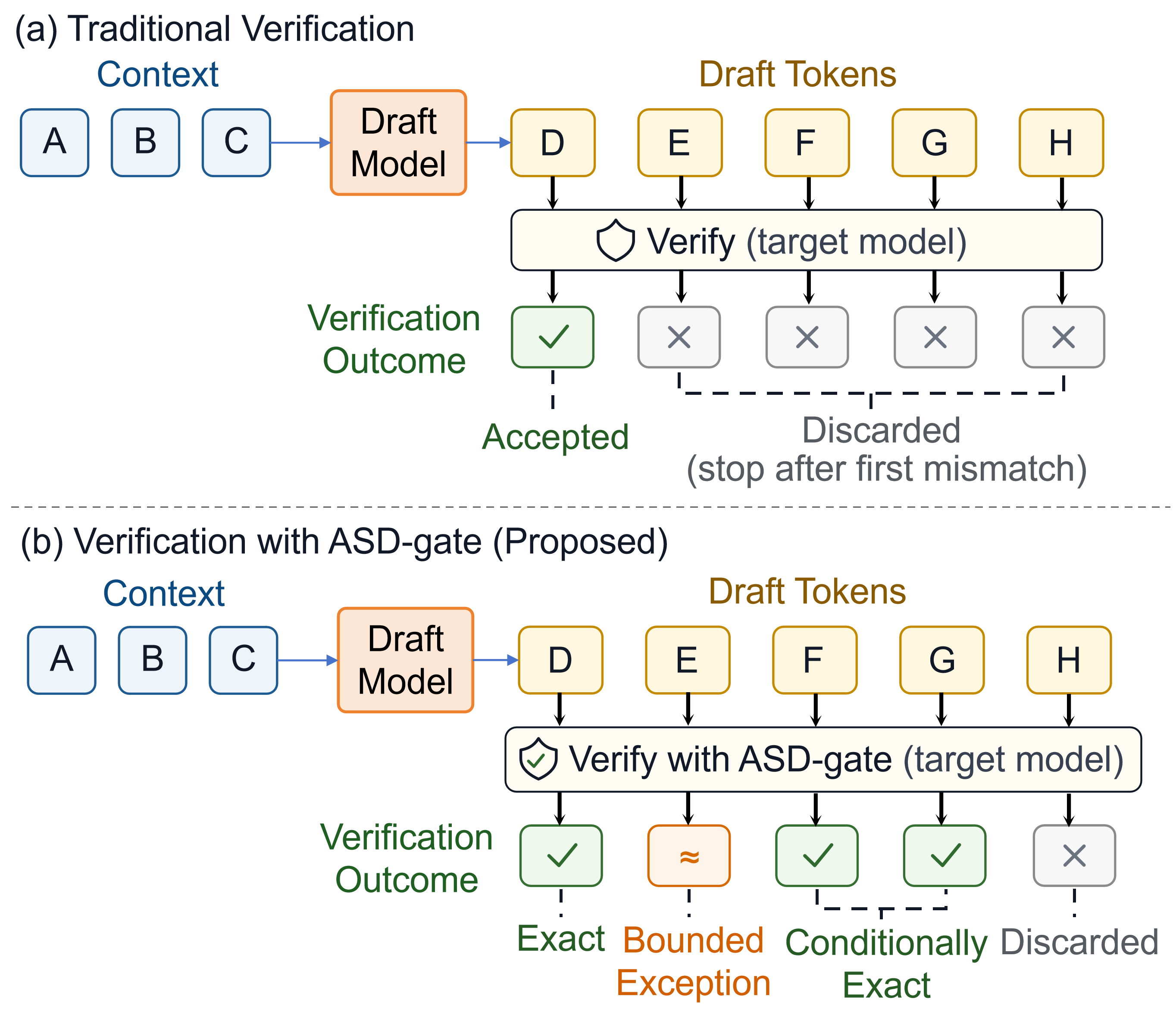}
\caption{\textbf{ASD verification.}
Strict verification stops at mismatch $E$. ASD accepts $E$ within budget, reuses the target-greedy suffix $F,G$ under the realized prefix, and recovers at $H$ after the same target pass.}
\label{fig:asd_overview}
\vspace{-0.4cm}
\end{figure}
\begin{figure*}[t]
\centering
\includegraphics[width=\textwidth]{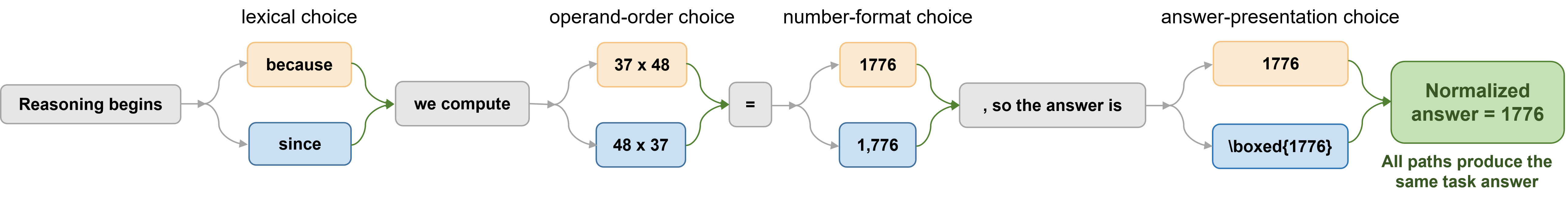}
\caption{\textbf{Different token paths can yield the same task-level answer.}
In this example, distinct reasoning trajectories produce the same normalized final answer. This motivates a separate task-quality audit, but does not imply that any approximate trajectory is individually exact or safe.}
\label{fig:overview}
\end{figure*}

The first mismatch also truncates a block that the target has already scored. In a teacher-forced verification pass, each later target row is conditioned on the preceding draft tokens, including any earlier mismatch. Consequently, if a verifier elects to accept an early mismatch, a subsequent contiguous draft suffix may already be target-greedy under the resulting realized prefix. These tokens can be committed without another target forward pass or additional approximate token decisions. Figure~\ref{fig:asd_overview} illustrates this opportunity: strict verification stops at mismatch $E$, whereas accepting $E$ enables reuse of $F$ and $G$, which are already target-greedy under the prefix containing $E$.
Figure~\ref{fig:overview} shows why token-level divergence does not necessarily imply task-level failure. Alternative connectives, commutative expressions such as $37\mathbin{\times}48$ and $48\mathbin{\times}37$, and answer formats such as ``1776,'' ``1,776,'' and $\backslash\mathrm{boxed}\{1776\}$ can preserve the same result. However, token mismatch is only an imperfect proxy for task quality and does not make an approximate trajectory exact or safe.

Accepting a non-greedy draft token therefore changes the decoding trajectory
and cannot be treated as a lossless optimization. The problem is not to ignore
mismatches, but to permit a small number of explicit exceptions while
controlling their cumulative magnitude over an entire request. A purely local
verifier may repeatedly accept individually inexpensive mismatches across
decoding rounds, allowing approximation to accumulate with output length.
Similarly, resetting an allowance in every draft block fails to account for
deviations already introduced earlier in the same completion.

In this paper, we introduce \textbf{Approximate Speculative Decoding (ASD)}, a training-free verifier that replaces first-mismatch truncation with budgeted longest-prefix selection. At each mismatch, ASD measures local target-logit regret and accepts the longest prefix satisfying a regret threshold, block-level exception cap, and request-level regret budget. It reuses subsequent draft tokens that remain target-greedy under the realized prefix and requires only the standard target verification pass. With a zero budget, ASD exactly recovers strict token-ID greedy verification. ASD bounds accumulated local regret but does not guarantee identical outputs, semantic preservation, or task correctness.
Our contributions are threefold:
\begin{itemize}
\item We formulate approximate greedy verification as a budgeted longest-contiguous-prefix selection problem, using a persistent request-level ledger to account for accepted draft--target mismatches.

\item We identify and formalize realized-prefix suffix reuse: after an accepted exception, a contiguous target-greedy suffix can be committed without additional approximate token decisions, provided that target and draft caches are updated consistently.

\item Experiments show that training-free verifier-side ASD improves fixed-workload throughput by up to $15.26\%$ over matched strict speculative decoding and by $7.78\%$ on average across seven Qwen3-14B + DSpark-14B tasks. It also improves acceptance on DeepSeek-V4-Flash (285B) with DSpark by roughly $10\%$--$16\%$ on GSM8K and MATH-500 under FP4-to-FP8 compatibility.
\end{itemize}

\section{Related Work}

\paragraph{Speculative decoding and proposal mechanisms.}
Speculative decoding accelerates autoregressive generation by using a lightweight drafter to propose tokens that are verified in parallel by a target model~\cite{leviathan2023fast,chen2023accelerating}. Prior work improves proposal quality and diversity through recurrent, retrieval-augmented, non-autoregressive, adaptive, multi-head, and self-speculative drafters~\cite{zhao2024ouroboros,cheng2024redrafter,cai2024medusa,li2024eagle,liu2024kangaroo,gao2026talon}. Tree-based methods further optimize the structure or width of candidates evaluated in one target pass~\cite{miao2024specinfer,chen2024sequoia,wang2025opttree,xiong2025dyspec,qin2025dynamic}. ASD is complementary: rather than changing proposal generation or organization, it changes the verifier's commitment decision after a proposed block has been scored.

\paragraph{Exact and efficient verification.}
Exact speculative sampling uses rejection-sampling corrections to preserve the target distribution, while greedy verification accepts draft tokens only when they match the target argmax~\cite{chen2023accelerating,leviathan2023fast}. Block Verification and Polybasic Speculative Decoding improve or analyze exact multi-token and multi-model verification~\cite{sun2025blockverification,wang2025polybasic}. Other methods reduce draft--target coordination or verifier cost through adaptive policies, overlapped drafting and verification, partial KV-state checks, or sparse target computation~\cite{liu2025adaptive,kumar2026speculative,tan2025specpv,wang2025sparseverification}. In contrast, ASD deliberately permits a bounded number of non-greedy draft tokens, targeting an explicit speed--behavior trade-off rather than exact target-output identity.

\paragraph{Relaxed verification.}
Recent relaxed methods improve acceptance using criteria beyond exact drafter--target agreement, including semantic judgments, divergence constraints, uncertainty-aware thresholds, and target-logit margins~\cite{bachmann2025judge,holsman2025fuzzy,sun2025ears,hao2026cactus,song2026mars}. ASD differs by formulating deterministic greedy verification as a longest \emph{contiguous} prefix-selection problem with a persistent request-level regret ledger. This ledger bounds cumulative target disagreement across decoding rounds, while suffix reuse commits subsequent target-greedy tokens without further approximate decisions or target-model forward passes.

\begin{figure}[!t]
\centering
\includegraphics[width=\linewidth]{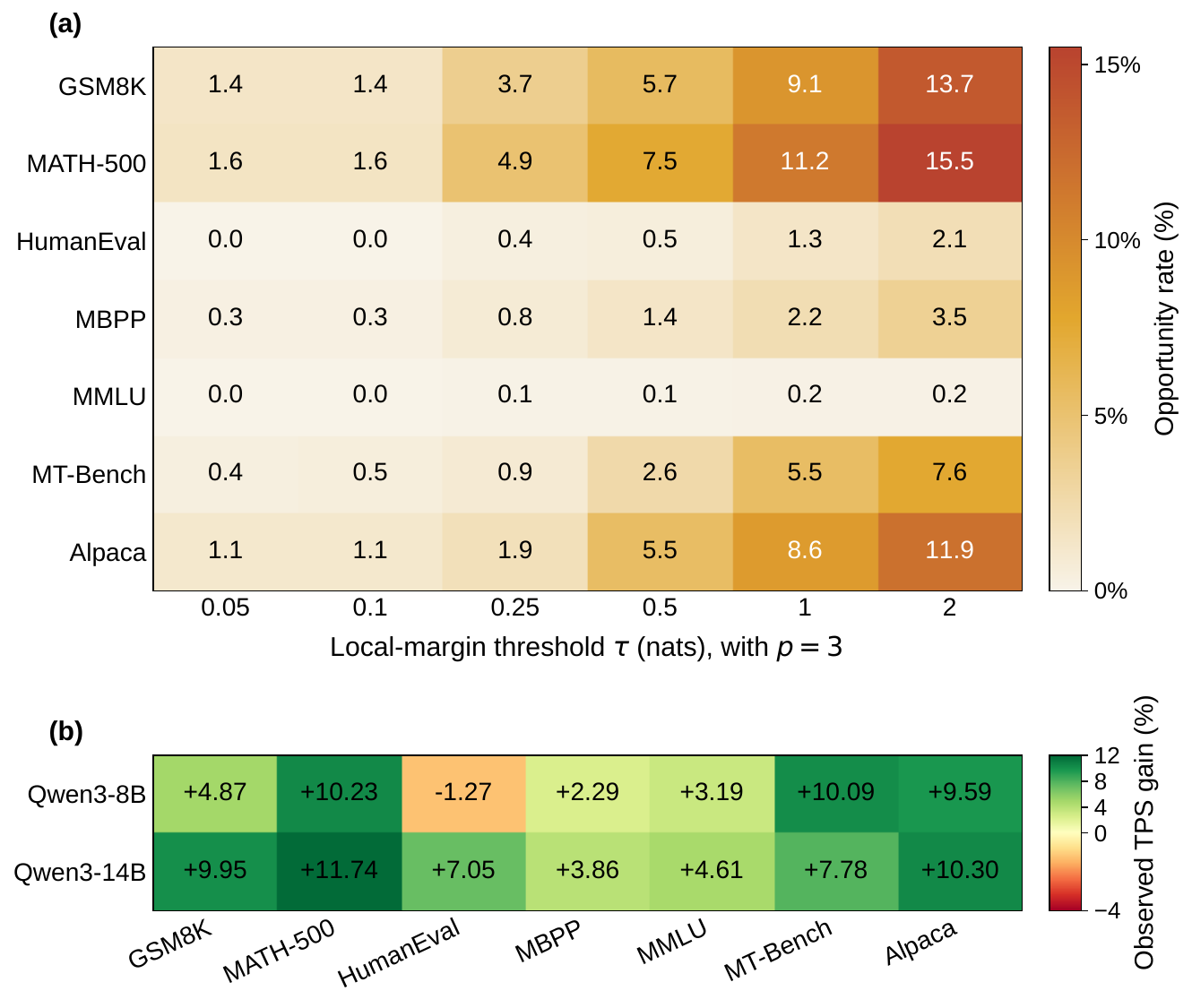}
\caption{\textbf{Verifier-side opportunity.}
(a) $\widehat{\Omega}(\tau,3)$ measures low-regret mismatches followed by at least three target-greedy draft tokens. (b) Fixed-workload TPS change over strict DSpark. The panels are complementary, not causal.}
\label{fig:opportunity}
\end{figure}

\section{Approximate Speculative Decoding}
\label{sec:asd}

\subsection{Verifier-side Opportunity}

Consider a committed history $h_0$. A draft model proposes a block $x_{1:K}=(x_1,\ldots,x_K)$, and a teacher-forced target pass produces logits
\begin{equation}
    z_i(\cdot) = \mathrm{Target}(h_0,x_{<i}), \qquad
    y_i^* = \arg\max_v z_i(v),
    \label{eq:target_rows}
\end{equation}
where $y_i^*$ denotes the target greedy token at position $i$. Standard greedy verification commits only the longest prefix that exactly matches the target greedy tokens:
\begin{equation}
    a_{\mathrm{strict}} =
    \max \left\{
        a \in \{0,\ldots,K\} :
        x_i = y_i^*,\ \forall i \leq a
    \right\}.
    \label{eq:strict_prefix}
\end{equation}
If $a_{\mathrm{strict}}<K$, it then appends the target recovery token $y_{a_{\mathrm{strict}}+1}^*$. This rule recovers the target greedy trajectory, but treats every mismatch identically regardless of how strongly the target prefers its argmax.

The first mismatch does not imply that all later target computation is useless. In particular, the target row $z_j$ for a later position $j$ is conditioned on the draft-forced prefix $(h_0,x_{<j})$, which includes any earlier draft mismatch. If that mismatch is accepted, this teacher-forced history becomes the realized decoding history. Thus, later draft tokens that equal their corresponding target argmax may already be target-greedy under the realized prefix. Figure~\ref{fig:asd_overview} illustrates this case: accepting $E$ makes the already scored tokens $F$ and $G$ reusable, whereas strict verification discards them after the mismatch at $E$.

To quantify the availability of this pattern, we define the local target-logit regret
\begin{equation}
    r_i = z_i(y_i^*) - z_i(x_i) \geq 0.
    \label{eq:local_regret}
\end{equation}
Because the softmax normalizer cancels, $r_i$ is also the target's conditional log-probability preference for $y_i^*$ over $x_i$. For a threshold $\tau$ and a desired reusable suffix length $p$, we measure the fraction of mismatch positions that are both low-regret and followed by $p$ contiguous target-greedy draft tokens:
\begin{equation}
\widehat{\Omega}(\tau,p)
=
\frac{1}{|\mathcal{M}|}
\sum_{(x,i)\in\mathcal{M}}
\mathbb{I}[r_i \leq \tau]
\mathbb{I}[i+p \leq K]
\prod_{\ell=i+1}^{i+p}
\mathbb{I}[x_\ell=y_\ell^*],
\label{eq:opportunity}
\end{equation}
where $\mathcal{M}$ is the set of draft--target mismatch positions over the evaluated blocks. Figure~\ref{fig:opportunity} shows that such potentially reusable suffixes occur across tasks and local-regret thresholds. The fixed-workload TPS changes in Figure~\ref{fig:opportunity}(b) provide complementary evidence that verifier-side reuse can translate into system-level gains, but do not establish a causal mapping from opportunity rate to throughput.
\begin{figure}[!t]
\centering
\includegraphics[width=\linewidth]{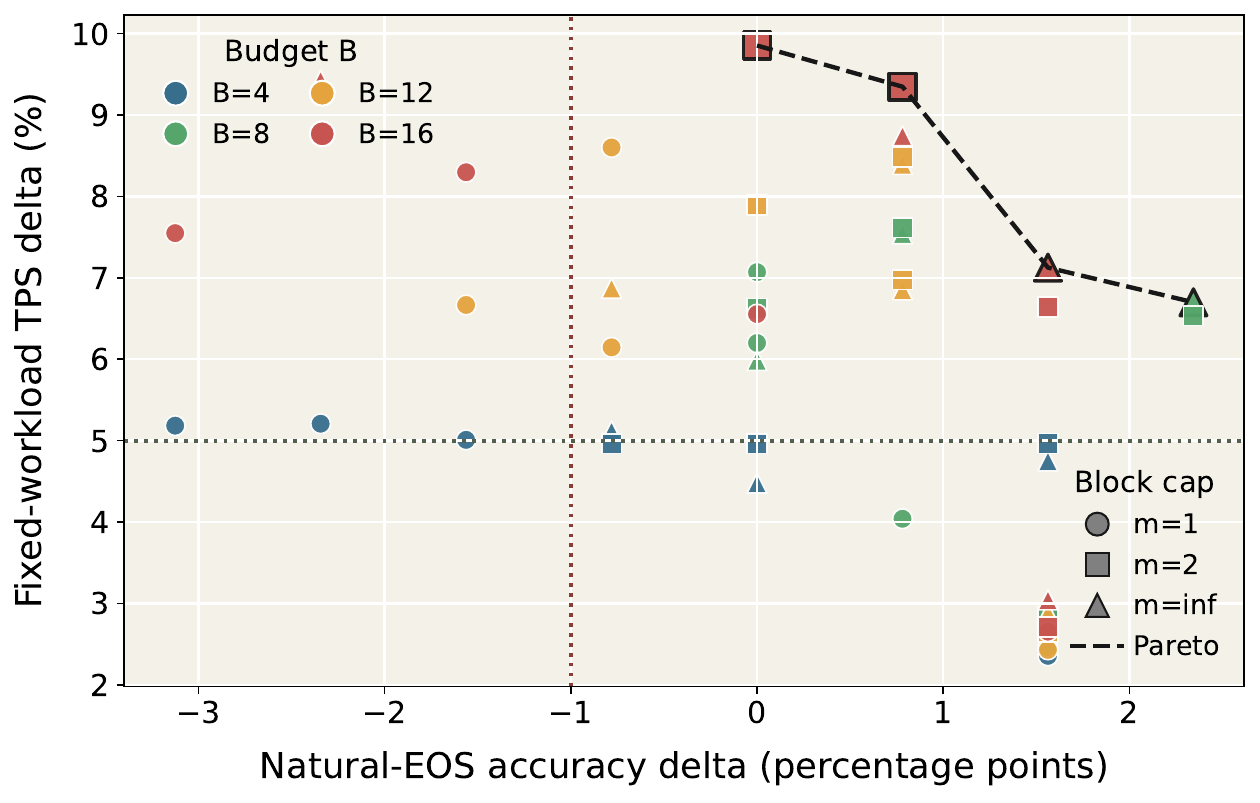}
\caption{\textbf{Preliminary speed--quality frontier.} Budget and cap move
the discovery trade-off, motivating joint local and request-level controls.
The frontier selects a conservative configuration; it is not a holdout result.}
\label{fig:tradeoff}
\end{figure}

These opportunities cannot be exploited by an unconstrained relaxed verifier. Accepting a non-greedy token changes the decoding trajectory, and individually inexpensive mismatches can accumulate across draft blocks and over long completions. Figure~\ref{fig:tradeoff} shows that the request budget and per-block exception cap move the throughput--accuracy trade-off, motivating separate controls over local, block-level, and request-level approximation. ASD implements these controls through a budgeted contiguous-prefix verifier.

\subsection{Budgeted Prefix Verification}

ASD treats a draft--target mismatch as an explicit exception. Let
\begin{equation}
    d_i = \mathbb{I}[x_i \neq y_i^*]
    \label{eq:mismatch_indicator}
\end{equation}
denote the mismatch indicator. For a mismatch at position $i$, the regret in Equation~\ref{eq:local_regret} measures the local target preference sacrificed by accepting $x_i$. It is an accounting quantity, rather than a calibrated error probability or a guarantee about downstream task quality.

Earlier exceptions can potentially unlock a longer already scored suffix. We therefore use the remaining proposal length
\begin{equation}
    q_i = K-i+1
    \label{eq:priority_proxy}
\end{equation}
as a training-free proxy for this opportunity. ASD applies a normalized local gate,
\begin{equation}
    \frac{r_i}{q_i} \leq g,
    \label{eq:local_gate}
\end{equation}
where $g$ controls the largest local regret accepted relative to the remaining proposal length. This proxy is not an optimality claim; it simply favors earlier exceptions only when their local cost is proportionate to their potential suffix-reuse opportunity.

Each request maintains a persistent state $(B,s)$, where $B$ is the total regret budget and $s$ is the regret already spent by accepted exceptions in previous decoding rounds. For a candidate prefix ending at position $t$, define its cumulative regret and exception count as
\begin{equation}
    C_t = \sum_{i=1}^{t} d_i r_i,
    \qquad
    N_t = \sum_{i=1}^{t} d_i.
    \label{eq:cumulative_state}
\end{equation}
Given a per-block exception cap $M$, the prefix $x_{1:t}$ is feasible if
\begin{align}
    C_t &\leq B-s, \label{eq:budget_constraint}\\
    N_t &\leq M, \label{eq:block_cap}\\
    \frac{r_i}{q_i} &\leq g
    \quad \text{for every mismatch } i\leq t. \label{eq:gate_constraint}
\end{align}
ASD returns the longest feasible prefix:
\begin{equation}
    a_{\mathrm{ASD}} =
    \max\left\{
        t \in \{0,\ldots,K\} :
        x_{1:t}\ \text{is feasible}
    \right\}.
    \label{eq:asd_prefix}
\end{equation}

The contiguous-prefix requirement is essential. ASD never skips an infeasible token to accept a later one: doing so would create a gap between the target rows used for verification and the history represented by the committed cache. The three constraints have distinct roles. The local gate filters individually costly exceptions, the block cap prevents exceptions from concentrating within one proposal block, and the persistent budget bounds their cumulative local regret throughout the request.
\begin{algorithm}[!t]
\caption{ASD for one greedy draft block}
\label{alg:asd}
\begin{algorithmic}[1]
\REQUIRE Draft block $x_{1:K}$; target greedy tokens $y_{1:K+1}^*$;
target logits $z_i(y_i^*)$ and $z_i(x_i)$ for $i\in[1,K]$;
state $(B,s)$; local gate $g$; block cap $M$
\IF{$B=0$}
    \RETURN exact greedy verifier output and unchanged state $(B,s)$
\ENDIF
\STATE $d_i \leftarrow \mathbb{I}[x_i \neq y_i^*]$ and
       $r_i \leftarrow z_i(y_i^*) - z_i(x_i)$ for $i\in[1,K]$
\STATE $q_i \leftarrow K-i+1$ for $i\in[1,K]$
\STATE $C_i \leftarrow \sum_{j=1}^{i} d_j r_j$ and
       $N_i \leftarrow \sum_{j=1}^{i} d_j$ for $i\in[1,K]$
\STATE Construct $f_{1:K}$ using Equation~\ref{eq:feasibility_mask}
\STATE $a \leftarrow$ length of the longest all-true prefix of $f_{1:K}$
\STATE $s \leftarrow s + \sum_{i=1}^{a} d_i r_i$
\IF{$a<K$}
    \RETURN $x_{1:a}$ followed by recovery token $y_{a+1}^*$,
    updated state $(B,s)$
\ELSE
    \RETURN $x_{1:K}$ followed by bonus token $y_{K+1}^*$,
    updated state $(B,s)$
\ENDIF
\end{algorithmic}
\end{algorithm}

\subsection{Decoding Procedure and Cache Semantics}

\paragraph{Suffix reuse under the realized prefix.} Suppose ASD commits $x_{1:a_{\mathrm{ASD}}}$. For every committed position $j \leq a_{\mathrm{ASD}}$, the target row $z_j$ was computed under the history $(h_0,x_{<j})$. Since ASD commits the same contiguous prefix, this is exactly the realized history before emitting $x_j$.

\begin{proposition}[Realized-prefix suffix reuse]
\label{prop:suffix_reuse}
If a committed draft token $x_j$ satisfies $x_j=y_j^*$, then $x_j$ is target-greedy under the realized ASD history and introduces no additional token-level exception.
\end{proposition}

The proposition does not restore equivalence to the strict target-greedy trajectory. An earlier accepted mismatch may already have changed the history on which subsequent target decisions are conditioned. Rather, it establishes that a contiguous suffix of target-greedy draft tokens following an accepted exception can be reused without spending further regret or executing another target forward pass.

\begin{table*}[t]
\centering
\small
\setlength{\tabcolsep}{6.0pt}
\renewcommand{\arraystretch}{1.15}

\begin{tabular}{
  @{}
  l
  S[table-format=1.2]
  S[table-format=1.2]
  >{\bfseries}S[table-format=1.2]
  S[table-format=+2.2(2)]
  S[table-format=1.2]
  >{\bfseries}S[table-format=1.2]
  S[table-format=+1.2]
  @{}
}
\toprule

\multirow{2}{*}{\textbf{Benchmark}}
&
\multicolumn{4}{c}{\textbf{Speedup}}
&
\multicolumn{2}{c}{\textbf{Threshold}}
&

\\

\cmidrule(lr){2-5}
\cmidrule(lr){6-7}

&
{\textbf{Target-only}}
&
{\textbf{Strict SD}}
&
{\textbf{ASD}}
&
{\textbf{ASD vs.\ Strict (\%)}}
&
{\textbf{Strict $\tau$}}
&
{\textbf{ASD $\tau$}}
&
{\textbf{ASD $\Delta$Acc.}}
\\

\midrule

GSM8K
& 1.00
& 5.58
& 6.30
& +10.08(50)
& 4.42
& 4.91
& +0.15
\\

MATH-500
& 1.00
& 6.88
& 7.32
& +11.73(38)
& 5.35
& 6.02
& +1.17
\\

HumanEval
& 1.00
& 4.68
& 5.14
& +6.90(61)
& 3.75
& 4.05
& -0.61
\\

MBPP
& 1.00
& 5.04
& 5.36
& +3.64(26)
& 3.97
& 4.15
& +0.54
\\

MMLU
& 1.00
& 3.30
& 3.54
& +4.41(57)
& 2.62
& 2.77
& +0.73
\\

MT-Bench
& 1.00
& 4.68
& 5.16
& +7.79(68)
& 3.62
& 3.92
& -0.64
\\

Alpaca
& 1.00
& 4.22
& 4.66
& +9.94(35)
& 3.24
& 3.60
& +0.10
\\

\bottomrule
\end{tabular}
\caption{Results for Qwen3-14B with DSpark-14B ($B=8$, $g=0.25$, $M=2$). Speedups are relative to target-only decoding. ASD consistently outperforms matched strict speculative decoding, with small task-dependent accuracy changes under natural-EOS decoding. Parentheses denote confidence-interval half-widths; accuracy changes are reported per task and not averaged across heterogeneous metrics.}
\label{tab:main-dspark14}
\end{table*}

\paragraph{Ledger update and recovery.} After committing the selected prefix, ASD updates the persistent ledger:
\begin{equation}
    s' = s + C_{a_{\mathrm{ASD}}}.
    \label{eq:ledger_update}
\end{equation}
By construction, Equation~\ref{eq:budget_constraint} ensures $s' \leq B$. Thus, the ledger bounds the sum of local regrets from all accepted exceptions along the realized request trajectory. It is not a complete-sequence likelihood-ratio bound, and it does not imply output identity, distributional equivalence, semantic preservation, safety, or task correctness.

If $a_{\mathrm{ASD}}<K$, the target row $z_{a_{\mathrm{ASD}}+1}$ already provides the recovery token $y_{a_{\mathrm{ASD}}+1}^*$ under the returned prefix. ASD emits this token and begins the next round from the updated history. If $a_{\mathrm{ASD}}=K$, ASD appends the bonus token $y_{K+1}^*$ from the final target row. In both cases, the target cache, draft cache, and request-level ledger must be advanced only along the emitted sequence. Proposal-only cache state beyond the returned prefix is discarded.

\paragraph{Vectorized implementation.} Algorithm~\ref{alg:asd} implements ASD after the ordinary target verification pass. Let
\begin{equation}
f_i =
(1-d_i)
\ \lor\
\left[
    \frac{r_i}{q_i}\leq g
    \ \land\
    C_i\leq B-s
    \ \land\
    N_i\leq M
\right]
\label{eq:feasibility_mask}
\end{equation}
be the per-position feasibility mask. The accepted length is the longest all-true prefix of $f_{1:K}$, which can be computed using cumulative sums followed by a cumulative product. The verifier therefore adds $O(K)$ arithmetic and temporary storage, but no target-model forward pass.

ASD requires only the target top token and top logit, together with the target logit assigned to each proposed draft token. In a tensor-parallel implementation, these values can be obtained without materializing an additional $K \times |\mathcal{V}|$ logit buffer. Finally, when $B=0$, ASD dispatches to ordinary token-ID greedy verification rather than accepting tokens solely because $r_i=0$. This strict fallback preserves standard greedy behavior even when multiple tokens tie for the maximum target logit.

\section{Experiments}

\subsection{Experimental Setup}

\paragraph{Models and runtime.}
We evaluate Qwen3-8B and Qwen3-14B targets with matched DSpark-8B-block7 and
DSpark-14B-block7 drafters, and test transfer to EAGLE3 with Llama-3.1-8B and
Qwen2.5-7B and Medusa with Llama-3.1-8B.  ASD is inserted only in the
verifier and uses the logits from the standard target verification pass; it
adds neither training nor target-model forward passes.  All comparisons use
greedy decoding, an unchanged target and prompt template, and identical
stop-token handling.  Unless stated otherwise, each task evaluation is
repeated four times, with each
strict--ASD--strict triplet run sequentially on one NVIDIA L20 GPU (46,068 MiB)
with Linux 5.4, Python 3.12.13, PyTorch
2.8.0+cu128, Transformers 4.55.2, and vLLM 0.11.0.

\paragraph{Tasks and evaluation protocol.}
We evaluate GSM8K, MATH-500, HumanEval, MBPP, MMLU, MT-Bench, and Alpaca.
Hyperparameter discovery uses disjoint GSM8K fixed-workload
($64\times256$ tokens) and natural-EOS (128 prompts, at most 512 tokens)
slices; selected controls are frozen before multi-task evaluation.  The full
configuration and per-run records are provided in the supplementary material.
Fixed-work decoding equalizes completion-token counts and measures ASD TPS
against the mean of adjacent strict roles; we retain only triplets with
matching strict hashes and token counts and at most $3\%$ baseline drift.
Reported TPS intervals are 95\% Student-$t$ half-widths in percentage points
of relative TPS gain.  Natural-EOS decoding separately audits accuracy,
length, and hash divergence, where hash changes diagnose altered trajectories
rather than task quality.

\subsection{Main Results}

We ask two questions: does relaxing only budget-feasible mismatches improve
end-to-end throughput over strict verification, and does the benefit persist
when the drafter family changes? 

Table~\ref{tab:main-dspark14} answers the
first question on Qwen3-14B with DSpark-14B using the frozen
$(B,g,M)=(8,0.25,2)$ configuration.  ASD improves fixed-workload TPS
on all seven tasks, by $3.64\%$--$11.73\%$ (mean $7.78\%$) over matched strict
SD.  The strongest gains occur on MATH-500 ($11.73\%$) and GSM8K ($10.08\%$),
where mean accepted length also rises by $0.67$ and $0.49$ tokens per round,
respectively.  Averaged over all tasks, $\tau$ increases from $3.85$ to $4.20$.
These aligned changes are the expected systems signature of ASD: a permitted
low-regret exception exposes a contiguous suffix that is already target-greedy
under the realized prefix, allowing the verifier to commit more tokens per
target pass.

\begin{table*}[t]
\centering
\small
\setlength{\tabcolsep}{3.8pt}
\renewcommand{\arraystretch}{1.15}

\begin{tabular}{
  l
  l
  l
  S[table-format=1.2]
  S[table-format=1.2]
  >{\bfseries}S[table-format=1.2]
  S[table-format=+2.2(2)]
  S[table-format=1.2]
  >{\bfseries}S[table-format=1.2]
  S[table-format=+1.2]
}
\toprule

\multirow{2}{*}{\textbf{Method}}
&
\multirow{2}{*}{\textbf{Target model}}
&
\multirow{2}{*}{\textbf{Benchmark}}
&
\multicolumn{4}{c}{\textbf{Speedup}}
&
\multicolumn{2}{c}{\textbf{Threshold}}
&

\\

\cmidrule(lr){4-7}
\cmidrule(lr){8-9}

&
&
&
{\textbf{Target-only}}
&
{\textbf{Strict SD}}
&
{\textbf{ASD}}
&
{\textbf{ASD vs.\ Strict(\%)}}
&
{\textbf{Strict $\tau$}}
&
{\textbf{ASD $\tau$}}
&
{\textbf{ASD $\Delta$Acc.}}
\\

\midrule

\multirow{4}{*}{\textit{DSpark}}
& Qwen3-14B & GSM8K
& 1.00 & 5.58 & 6.30 & +10.08(50) & 4.42 & 4.91 & +0.15 \\

& Qwen3-14B & MMLU
& 1.00 & 3.30 & 3.54 & +4.41(57) & 2.62 & 2.77 & +0.22 \\

& Qwen3-8B & GSM8K
& 1.00 & 5.86 & 6.16 & +4.88(43) & 5.31 & 5.62 & -0.15 \\

& Qwen3-8B & MMLU
& 1.00 & 2.78 & 2.96 & +3.05(90) & 2.55 & 2.67 & +0.56 \\

\addlinespace[3pt]
\midrule
\addlinespace[1pt]

\multirow{4}{*}{\textit{EAGLE3}}
& Llama-3.1-8B & GSM8K
& 1.00 & 4.74 & 5.34 & +15.26(15) & 3.36 & 3.88 & -1.17 \\

& Llama-3.1-8B & MMLU
& 1.00 & 2.88 & 3.22 & +11.94(30) & 2.10 & 2.37 & +0.00 \\

& Qwen2.5-7B & GSM8K
& 1.00 & 4.10 & 4.30 & +6.08(30) & 2.86 & 3.03 & -0.39 \\

& Qwen2.5-7B & MMLU
& 1.00 & 2.50 & 2.66 & +8.11(21) & 1.76 & 1.91 & +0.50 \\

\addlinespace[3pt]
\midrule
\addlinespace[1pt]

\multirow{2}{*}{\textit{Medusa}}
& Llama-3.1-8B & GSM8K
& 1.00 & 2.70 & 2.94 & +6.72(13) & 2.03 & 2.17 & -1.52 \\

& Llama-3.1-8B & MMLU
& 1.00 & 1.82 & 1.94 & +4.70(14) & 1.37 & 1.44 & +0.00 \\

\bottomrule
\end{tabular}
\caption{
Cross-family ASD results under matched fixed-work and natural-EOS
protocols. Speedups are measured relative to target-only decoding.
Values in parentheses denote the corresponding confidence intervals.
}
\label{tab:cross-family}
\end{table*}

\begin{table}[t]
\centering
\small
\setlength{\tabcolsep}{6.0pt}
\renewcommand{\arraystretch}{0.98}
\begin{tabular}{@{}clcc@{}}
\toprule
\textbf{Control} & \textbf{Value} &
\textbf{Fixed-work gain (\%)} &
\textbf{$\Delta$Acc. (pp)} \\
\midrule
\multirow{4}{*}{$B$}
& 4  & 4.96 & $-0.78$ \\
& \cellcolor{blue!12}8
& \cellcolor{blue!12}6.53
& \cellcolor{blue!12}$+2.34$ \\
& 12 & \textbf{6.98} & $+0.78$ \\
& 16 & 6.64 & $+1.56$ \\
\midrule

\multirow{4}{*}{$g$}
& 0.125 & 2.84 & $+1.56$ \\
& \cellcolor{blue!12}0.25
& \cellcolor{blue!12}6.53
& \cellcolor{blue!12}$+2.34$ \\
& 0.5 & \textbf{7.62} & $+0.78$ \\
& $\infty$ & 6.62 & $+0.00$ \\
\midrule

\multirow{3}{*}{$M$}
& 1 & 6.20 & $+0.00$ \\
& \cellcolor{blue!12}2
& \cellcolor{blue!12}6.53
& \cellcolor{blue!12}$+2.34$ \\
& $\infty$ & \textbf{6.70} & $+2.34$ \\
\bottomrule
\end{tabular}
\caption{One-at-a-time ASD control sweeps on GSM8K. Blue cells indicate the selected operating points.}
\label{tab:ablation-response}
\end{table}

\begin{figure}[!t]
\centering
\includegraphics[width=\linewidth]{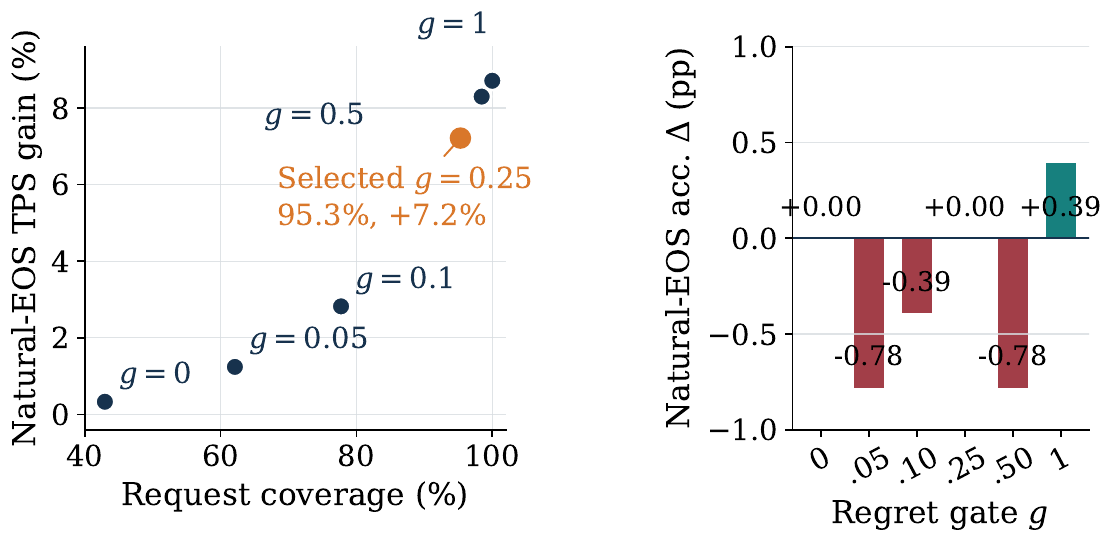}
\caption{Regret-gate sweep on GSM8K under natural-EOS decoding ($n=256$, one seed).}
\label{fig:regret_gate_tradeoff}
\end{figure}

The natural-EOS audit separates this throughput result from behavioral
preservation.  On the primary DSpark-14B matrix, five of seven task rows have
non-negative accuracy changes, while the two observed reductions are small
($-0.61$ points on HumanEval and $-0.64$ on MT-Bench).  At the same time,
GSM8K and MATH-500 exhibit hash divergence above $95\%$, even though their
measured accuracy does not decrease.  Thus, the result is not an
output-preserving speedup: the accuracy audit shows that the selected budget
can retain task performance on most measured tasks, and the hash audit makes
the altered trajectories explicit.

Table~\ref{tab:cross-family} tests generalization across DSpark, EAGLE3, and
Medusa.  ASD improves throughput in every reported cell, spanning
$3.05\%$--$15.26\%$ with a $7.52\%$ mean gain across the ten cells.  The
accepted length increases in each cell as well ($+0.07$ to $+0.52$ tokens),
which ties the cross-family gains to the same verifier-side mechanism rather
than a family-specific change in proposal generation.  In particular, the
largest gains, $15.26\%$ on Llama-3.1-8B + EAGLE3 for GSM8K and $11.94\%$ for
MMLU, show that budgeted prefix selection remains effective with a different
draft architecture.

Taken together, the two tables show that ASD's throughput benefit is systematic rather than driven by a small subset of favorable workloads.  Gains are positive on all seven tasks in the primary DSpark-14B evaluation and in all ten cells of the cross-family study; moreover, every reported 95\% confidence interval remains strictly above zero.  These improvements are obtained on top of already strong strict-SD baselines, which span $1.82\times$--$6.88\times$ target-only speedup, with ASD raising the corresponding range to $1.94\times$--$7.32\times$.

The benefit also covers a broad range of baseline acceptance levels.  Across the two tables, strict accepted length ranges from $1.37$ to $5.35$ tokens per round, yet ASD increases it in every setting.  This consistency across four target models and three drafter families indicates that the verifier does not depend on a particular proposal architecture or a narrow baseline-acceptance regime.  Because ASD operates only after ordinary target scoring, these gains require neither retraining the drafter nor an additional target-model forward pass, making the method complementary to improvements in proposal generation.

The accuracy columns nevertheless reinforce that ASD should be viewed as a controlled approximation rather than an output-preserving optimization.  Five of seven primary-task rows and six of ten cross-family cells show non-negative measured accuracy changes, while the remaining reductions are task dependent and no larger than $1.52$ percentage points.  Thus, throughput improvement is considerably more consistent than the direction of the quality change. Together with the observed hash divergence, this asymmetry supports ASD's intended deployment model: freeze the verifier controls on disjoint data, then audit natural-EOS task quality for each target, drafter, and workload.


\subsection{Analysis and Discussion}

The ablations test whether ASD's improvement follows the behavior predicted by budgeted prefix verification rather than reflecting a fixed increase in acceptance. Table~\ref{tab:ablation-response} varies the request budget $B$, local gate $g$, and per-block cap $M$ around the frozen $(8,0.25,2)$ configuration. These controls address complementary failure modes: $B$ bounds cumulative request-level deviation, $g$ rejects locally costly mismatches, and $M$ prevents exceptions from concentrating within a draft block. In each sweep, the other two controls remain fixed; boldface marks the highest fixed-work gain, and blue shading indicates the frozen configuration used in the main results. Reporting fixed-work gain together with the separate natural-EOS accuracy audit exposes the operating trade-off. The selected configuration therefore balances throughput gains against the measured task behavior used for selection.

\begin{figure}[t]
\centering
\includegraphics[width=\linewidth]{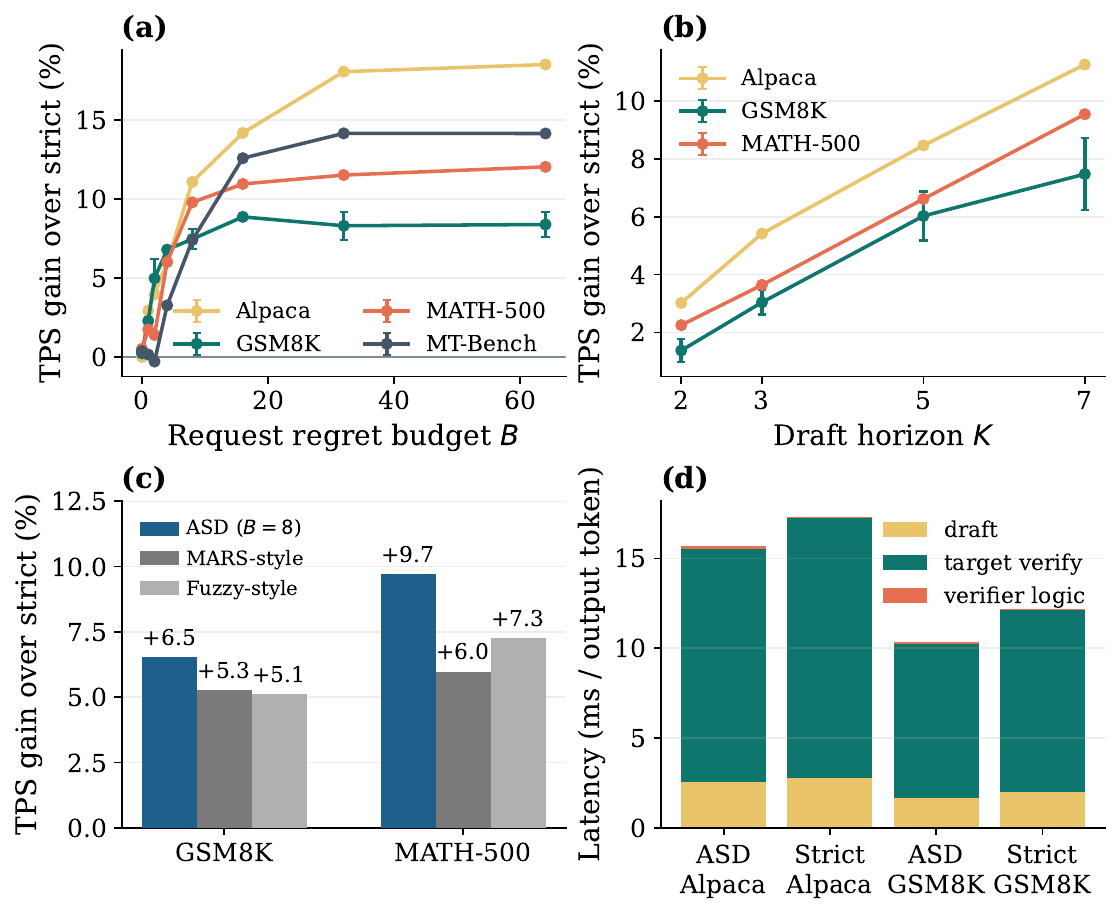}
\caption{ASD ablations with Qwen3-14B + DSpark-14B: (a) request-budget response, (b) draft-horizon response under fixed-work decoding ($n=64$, one seed), (c) MARS-style and Fuzzy-style local controls at $B=8$, and (d) synchronized per-token latency.}
\label{fig:finegrained-ablation}
\end{figure}

\paragraph{Budget control and reusable suffix opportunity.}
Figure~\ref{fig:finegrained-ablation}(a) makes the request budget an
interpretable operating knob.  At $B=0$, ASD's verifier is exactly strict
greedy verification.  Increasing $B$ admits low-regret exceptions and yields
substantial throughput gains on Alpaca, GSM8K, MATH-500, and MT-Bench; the
flattening at larger budgets shows that the useful suffix opportunity is finite
rather than an artifact of unconstrained acceptance.  The selected $B=8$
therefore lies on the rising part of this response, before the high-budget
plateau.  Figure~\ref{fig:finegrained-ablation}(b) independently varies the
draft horizon under fixed-work decoding ($n=64$, one seed).  ASD outperforms
strict verification at every tested horizon, and the gain increases with $K$
on all three workloads.  This coupled horizon response is the expected
signature of realized-prefix suffix reuse: more target-scored positions remain
available after an early draft mismatch.

\begin{figure}[t]
\centering
\includegraphics[width=\linewidth]{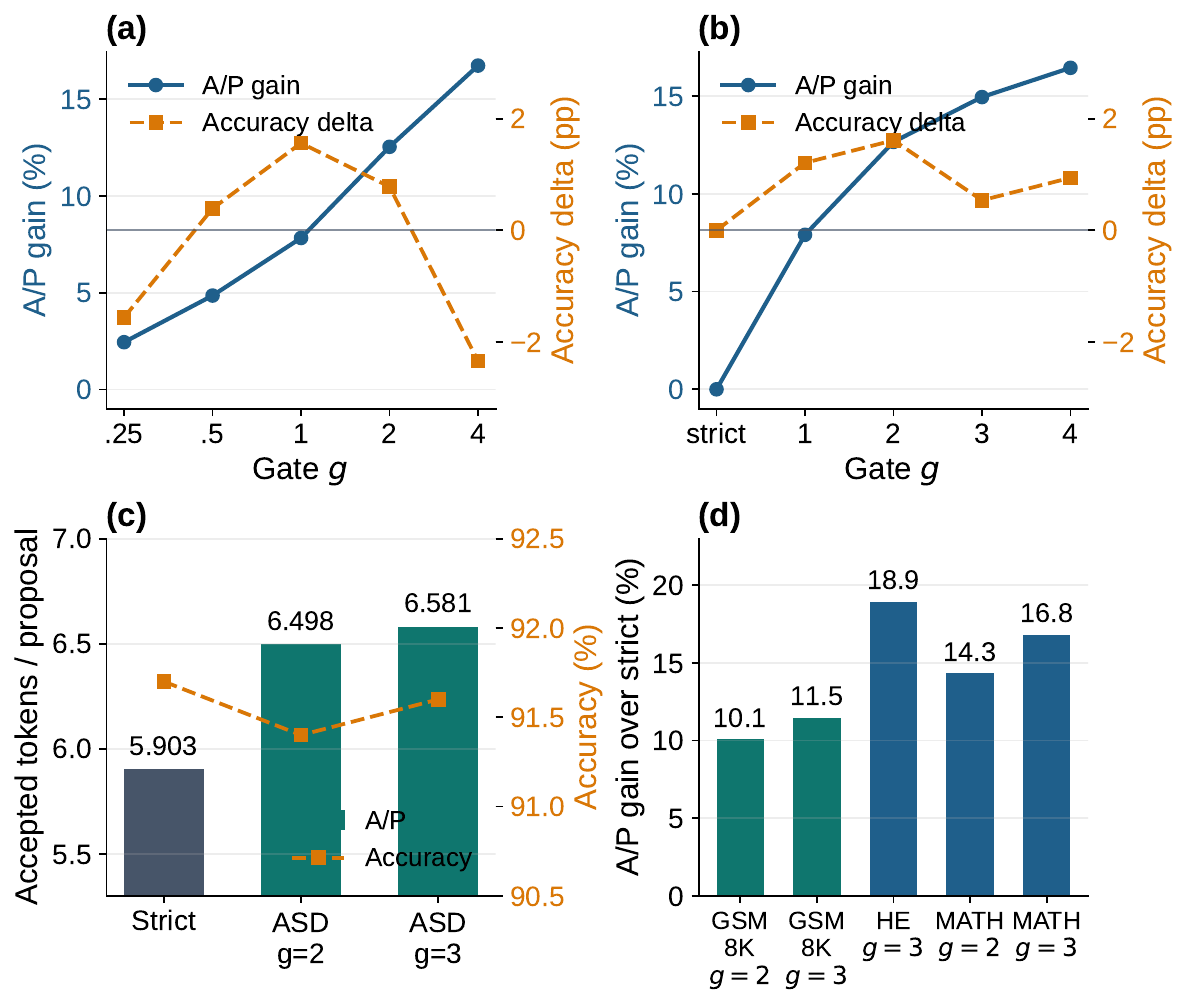}
\caption{DeepSeek-V4-Flash + DSpark acceptance ablations on eight H20 GPUs: (a) GSM8K gate sweep ($n=256$), (b) GSM8K power-set evaluation ($n=744$), (c) GSM8K-Confirm validation ($n=1{,}000$), and (d) cross-task accepted tokens per proposal (A/P).}
\label{fig:deepseek-v4-ablation}
\end{figure}

\paragraph{Budget-matched verifier controls and latency.}
Figure~\ref{fig:finegrained-ablation}(c) compares ASD with audited
MARS-style and Fuzzy-style local verifier controls after applying the same
maximum-realized-regret criterion.  ASD is best on both tasks: $+6.5\%$ on
GSM8K, versus $+5.3\%$ and $+5.1\%$, and $+9.7\%$ on MATH-500, versus
$+6.0\%$ and $+7.3\%$, respectively.  This comparison distinguishes ASD's
request-level accounting from a locally permissive rule that has no persistent
budget.  Figure~\ref{fig:finegrained-ablation}(d) closes the systems loop.
In the synchronized profiling shown, verifier logic increases by only $0.083$
ms per output token on Alpaca and $0.045$ ms on GSM8K, while target
verification falls by $1.479$ and $1.509$ ms, respectively.  Thus, the gain
is not created by hiding verifier overhead: the bounded exceptions commit
useful target-scored suffixes and reduce expensive verifier rounds.

\paragraph{The regret gate controls the speed--behavior frontier.}
Figure~\ref{fig:regret_gate_tradeoff} isolates $g$ under natural-EOS decoding.
This GSM8K audit uses $n=256$ prompts and one seed.
Relaxing the gate increases both request coverage and throughput gain, as more
near-tied draft tokens become eligible exceptions.  The selected $g=0.25$
covers $95.3\%$ of requests and yields a $7.2\%$ TPS gain without a measured
GSM8K accuracy change.  More permissive gates produce small accuracy
fluctuations, demonstrating why local target disagreement must be gated
rather than treating every mismatch as reusable.  Collectively, these
ablations connect the controls to ASD's intended mechanism: a bounded,
low-regret exception unlocks subsequent target-greedy suffix tokens and
thereby reduces the number of verifier rounds.

\paragraph{Large-model acceptance ablation.}
Figure~\ref{fig:deepseek-v4-ablation} summarizes the development, confirmation, and fresh-validation acceptance experiments; complete numerical results and per-run records are provided in the supplementary material. In the development GSM8K sweep, relaxing $g$ increases accepted draft tokens per proposal (A/P), although accuracy fluctuations motivate confirmation. The 744-example GSM8K power-set evaluation preserves this trend. On the independent 1,000-example GSM8K-Confirm split, $g\in\{2,3\}$ improves A/P by $10.08\%$--$11.48\%$ while changing accuracy by at most $0.30$ percentage points. HumanEval and MATH-500 show similar acceptance gains. DSpark uses FP4 mixed precision for DeepSeek-V4-Flash, whereas the Hopper-based H20 supports native FP8 rather than FP4; the resulting compatibility path inserts additional quantization--dequantization (Q/DQ) operations, so these experiments therefore characterize task accuracy and verifier-side acceptance only.



\section{Conclusion}

ASD is a training-free, verifier-side method that uses bounded low-regret
exceptions to reuse target-scored suffix tokens.  It reaches up to $15.26\%$
throughput improvement over strict verification, with a $7.78\%$ average gain
on seven Qwen3-14B + DSpark-14B tasks and positive gains in every reported
DSpark, EAGLE3, and Medusa setting.  By requiring no target or drafter
training, ASD provides a broadly compatible path to faster greedy blockwise
speculative decoding, subject to task-level quality auditing.  On
DeepSeek-V4-Flash with DSpark, ASD also raises verifier-side acceptance
across additional datasets, including roughly $10\%$--$16\%$ gains on GSM8K
and MATH-500. 
\clearpage
\bibliography{references}


\end{document}